%% file: example_paper.tex
\documentclass{article}

\usepackage{microtype}
\usepackage{graphicx}
\usepackage{subcaption}
\usepackage{booktabs} % for professional tables

\usepackage{xurl}
\usepackage{hyperref}

\usepackage[preprint]{icml2026}

\usepackage{amsmath}
\usepackage{amssymb}
\usepackage{mathtools}
\usepackage{amsthm}

\usepackage{diagbox}
\usepackage{bm}
\usepackage{tabularx}

\usepackage[capitalize,noabbrev]{cleveref}

\theoremstyle{plain}

\theoremstyle{definition}

\theoremstyle{remark}

\usepackage[textsize=tiny]{todonotes}

\icmltitlerunning{You Only Forward Once: An Efficient Compositional Judging Paradigm}

\begin{document}

\twocolumn[
  \icmltitle{You Only Forward Once: An Efficient Compositional Judging Paradigm}



  \icmlsetsymbol{equal}{*}

  \begin{icmlauthorlist}
    \icmlauthor{Tianlong Zhang}{equal,hit}
    \icmlauthor{Hongwei Xue}{ali}
    \icmlauthor{Shilin Yan}{ali}
    \icmlauthor{Di Wu}{ali}
    \icmlauthor{Chen Xu}{ali}
    \icmlauthor{Guannan Zhang}{ali}
    \icmlauthor{Yunyun Yang}{hit}
  \end{icmlauthorlist}

  \icmlaffiliation{hit}{School of Science, Harbin Institute of Technology, Shenzhen, Guangdong Province, China}
  \icmlaffiliation{ali}{Accio, Alibaba Group, Hangzhou, Zhejiang Province, China}

  \icmlcorrespondingauthor{Yunyun Yang}{yangyunyun@hit.edu.cn}
  \icmlcorrespondingauthor{Tianlong Zhang}{24b958007@stu.hit.edu.cn}

  \icmlkeywords{Multimodal Large Language Models, Cross-modal Retrieval, MLLM-as-Judge}

  \vskip 0.3in
]



\printAffiliationsAndNotice{}  % no special notice (required even if empty)

\input{sec/0_abstract}

\input{sec/1_intro}
\input{sec/2_related_works}
\input{sec/3_method}

\input{sec/4_experiments}

\input{sec/5_conclusions}

\bibliography{example_paper}
\bibliographystyle{icml2026}

%%%%%%%%%%%%%%%%%%%%%%%%%%%%%%%%%%%%%%%%%%%%%%%%%%%%%%%%%%%%%%%%%%%%%%%%%%%%%%%
%%%%%%%%%%%%%%%%%%%%%%%%%%%%%%%%%%%%%%%%%%%%%%%%%%%%%%%%%%%%%%%%%%%%%%%%%%%%%%%
% APPENDIX

\end{document}

%% file: sec/0_abstract.tex
\begin{abstract}
Multimodal large language models (MLLMs) show strong potential as judges. However, existing approaches face a fundamental trade-off: adapting MLLMs to output a single score misaligns with the generative nature of MLLMs and limits fine-grained requirement understanding, whereas autoregressively generating judging analyses is prohibitively slow in high-throughput settings. Observing that judgment reduces to verifying whether inputs satisfy a set of structured requirements, we propose \textbf{YOFO}, a template-conditioned method that judges all requirements in a single forward pass. Built on an autoregressive model, YOFO accepts a structured requirement template and, in one inference step, produces a binary yes/no decision for each requirement by reading the logits of the final token associated with that requirement. This design yields orders-of-magnitude speedups while preserving interpretability. 
Extensive experiments show that YOFO not only achieves state-of-the-art results on standard recommendation datasets, but also supports dependency-aware analysis---where subsequent judgments are conditioned on previous ones---and further benefits from post-hoc CoT.
% Extensive experiments demonstrate that YOFO achieves state-of-the-art performance on recommendation datasets, enables dependency-aware analysis in which later judgements condition on earlier results, and benefits from post-hoc CoT.
\end{abstract}

%% file: sec/1_intro.tex
\section{Introduction}
\label{sec:intro}

\begin{figure}[t]
    \centering
    \includegraphics[width=1\linewidth]{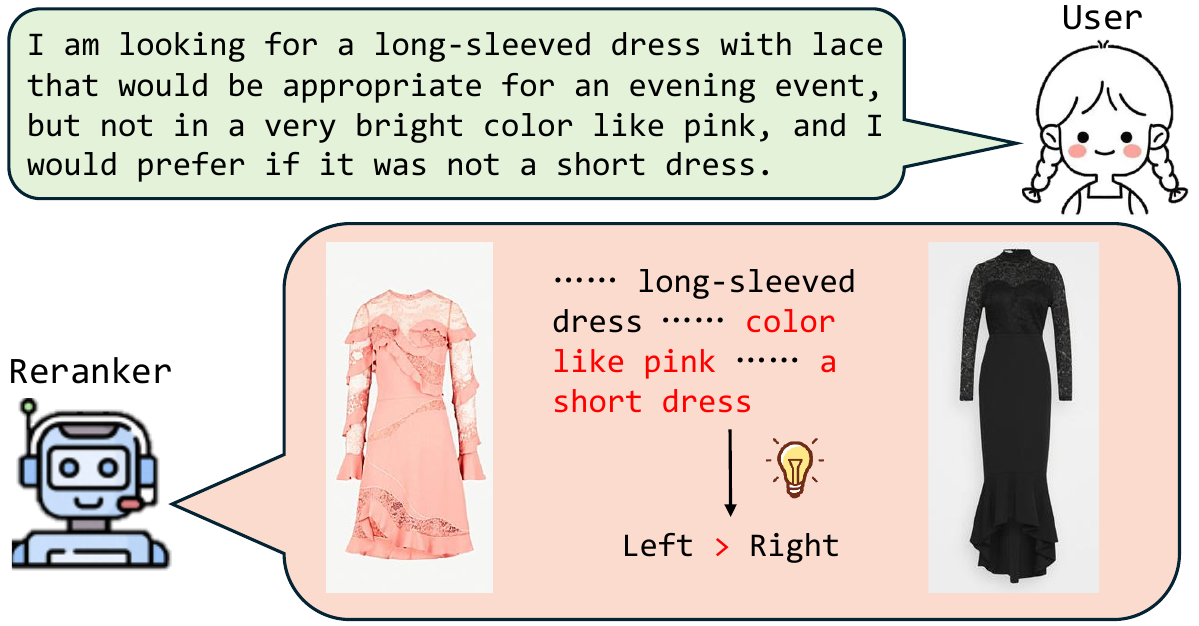}
    \caption{An example illustrating \textbf{a limitation of Jina-Reranker-M0}. The user seeks a midi or long dress in a non-pink color for an evening event, yet the model ranks a short pink dress higher than a long black dress that is well-suited for such an occasion.}
    \label{fig:jina}
    % \vspace{-6mm}
\end{figure}

Large language models (LLMs) \cite{team2024qwen2, yang2025qwen3, ahmed2025qwen, bai2023qwen, liu2024deepseek, guo2025deepseek} have achieved remarkable progress, and advances in multimodal learning now enable them to process diverse modalities. Multimodal Large Language Models (MLLMs) \cite{bai2025qwen2, wang2024qwen2, bai2023qwenvl, lu2025ovis2, lu2024ovis} have recently demonstrated significant promise in information matching due to their ability to jointly process images, videos, and natural language queries, making them far more applicable in real-world scenarios than traditional unimodal retrievers. In particular, state-of-the-art MLLM-based rerankers harness powerful vision encoders and causal attention mechanisms to effectively fuse visual and textual information, usually ultimately regressing a single relevance score, and achieves strong performance in cross-modal matching tasks such as image search and e-commerce recommendation. However, despite these advances, outputting only a scalar relevance score leads to substantial information loss, as fine-grained semantic distinctions in complex user queries—such as compositional intent, attribute-level preferences, or contextual references—are inevitably collapsed into a single value. As an example, \cref{fig:jina} illustrates a recommendation failure of the state-of-the-art (SOTA) multimodal reranker, Jina-Reranker-M0 \cite{jina2025blog}, which prioritizes superficial semantic overlap between the user’s query and the pink dress, resulting in an inappropriately high ranking for this item. Therefore, achieving accurate and fine-grained understanding of intricate multimodal queries has become an urgent challenge that must be addressed to realize the full potential of next-generation information matching systems.

\begin{figure*}[t]
    \centering
    \includegraphics[width=\linewidth]{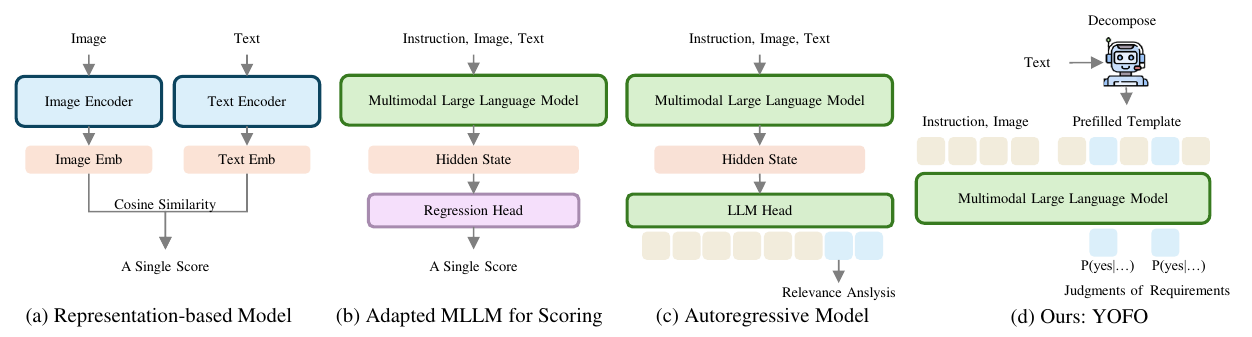}
    \caption{\textbf{Comparisons on Different Information Matching Methods.} (a) embeds inputs and then calculate similarity. (b) adapts an MLLM to output a single relevance score. (c) reformulates the matching problem as an autoregressive task, outputing relevance analysis. (d) Our method first decompose the text into a set of fundamental requirements and prefills them into a predefined template. Then the MLLM takes the template and the image as input and output whether each requirement is satisfied after a single forward pass.}
    \label{fig:prev}
    % \vspace{-4mm}
\end{figure*}

Recent studies on matching can be grouped into four categories. \textit{1)} Representation-based architectures, as illustrated in \cref{fig:prev} (a), independently embed the query and the document and then compute their similarity \cite{radford2021learning, huang2013learning, gao2014modeling, shen2014learning, palangi2016deep,wan2016deep,cohen2016end}. However, fixed-size embeddings limit these models’ capacity, making it difficult to capture all relevant information and to meet fine-grained requirements \cite{xu2025survey}. \textit{2)} Interaction-based models have emerged to mitigate these limitations: they first compute low-level interaction representations between the query and document, and then a neural network analyzes these patterns to output a scalar relevance score for the query–document pair \cite{hu2014convolutional, dai2019deeper, nogueira2019multi, gao2022long}. These approaches achieve higher ranking quality than representation-based methods but are computationally more expensive and remain constrained by representational dimensionality. \textit{3)} MLLM/LLM-based architectures,  adapted to output a single relevance score improve matching quality by leveraging the deep semantic understanding and long-context capacity of large language models \cite{ma2024fine}, as depicted in \cref{fig:prev} (b). However, this design is misaligned with the native generative objective of LLMs and undermines fine-grained input understanding. \textit{4)} LLMs and MLLMs can serve as native judges in a paradigm termed generative matching \cite{yoon2024listt5, yang2025rank}, which reframes the task as autoregressive generation aligned with their generative nature, as demonstrated in \cref{fig:prev} (c). Nevertheless, these methods incur high computational overhead due to token-by-token decoding, making them difficult to deploy in real-world applications.

To address this dilemma, we observe that information matching essentially reduces to verifying whether a document satisfies a set of input requirements. We therefore propose \textbf{YOFO}, a single-forward-pass, template-conditioned judging paradigm that simultaneously determines, for each requirement in a predefined template, whether it is satisfied (yes/no). Specifically, YOFO adapts an autoregressive model to accept a structured requirement template and, after a single forward pass, reads the logits of the final token associated with each requirement; the induced next-token distribution discriminates between yes and no, enabling binary decisions without autoregressive decoding. Consequently, the model produces all requirement judgments in one forward pass, improving throughput while preserving interpretability. During training, YOFO follows standard next-token prediction but applies supervision only to positions that predict yes/no or reasoning tokens, encouraging precise and well-grounded judgments. At inference time, we first obtain a template by prompting an LLM to decompose the user’s query into a set of requirements formatted according to a predefined schema (e.g., for “a blue, hooded, long-sleeve top without a chest logo,” the requirements are: blue, hooded, long-sleeved, no chest logo). YOFO then consumes the template and, in a single forward pass, reads the final-token logits for each requirement and outputs a yes/no decision by applying softmax and selecting the higher-probability token between ``yes'' and ``no''. The resulting judgments can be post-processed as needed—for example, mapped to a relevance score using a human-defined mapping.

To evaluate the efficacy of our method, we derive a training set from SA-1B \cite{kirillov2023segment} and train the model on it to improve generalizability. We then construct a test set from LRVS-Fashion \cite{lepage2023lrvs} to assess recommendation performance in a different domain. Extensive experiments demonstrate that our method accurately produces judgments conditioned on the input and template, and that later judgments can condition on earlier ones, enabling dependency-aware analysis. Furthermore, YOFO also benefits from post-hoc Chain-of-Thought (CoT), which encourages the model to reason implicitly before judging.

In summary, our contributions are as follows: (1) We introduce YOFO, an efficient compositional judging paradigm that judges a series of requirements concurrently in a single forward pass. (2) We demonstrate that YOFO enables dependency-aware analysis by conditioning later judgments on earlier ones. (3) We show that YOFO benefits from post-hoc CoT. (4) We further show that YOFO achieves state-of-the-art performance on recommendation tasks compared with traditional methods that output a single relevance score, while preserving high throughput.

%% file: sec/2_related_works.tex
\section{Related Works}
\label{sec:related_works}

\subsection{Multimodal Large Language Models}
\label{sebsec:mllm}

Multimodal Large Language Models (MLLMs) \cite{bai2025qwen2, wang2024qwen2, bai2023qwenvl, lu2025ovis2, lu2024ovis, hong2025worldsense, ma2024visual, ma2024multi, xue2023clip} have significantly extended the capabilities of Large Language Models (LLMs) \cite{team2024qwen2, yang2025qwen3, ahmed2025qwen, bai2023qwen, liu2024deepseek, guo2025deepseek, li2025adaptive} by enabling them to perceive, reason over, and generate responses grounded in multimodal inputs. However, the dense visual token sequences required for such alignment incur prohibitive computational costs. Recent approaches address this by either compressing tokens while preserving fidelity or rethinking how visual information is integrated. For instance, CrossLMM \cite{yan2025crosslmm} employs dual cross-attention mechanisms to drastically reduce token count with minimal performance degradation. Similarly, MM-GEM \cite{ma2024multi} uses a PoolAggregator to support both generative and embedding tasks. These advances demonstrate that strategic token reduction can maintain multimodal performance at lower cost. 
\subsection{Rerankers}
\label{subsec:rerankers}

Reranking serves as a critical refinement stage in modern retrieval systems, aiming to re-order an initial candidate set by modeling fine-grained query–document interactions. Early neural rerankers employed representation-based architectures, calculating embeddings independently and then similarity, such as DSSM \cite{huang2013learning, gao2014modeling}, Convolutional DSSM \cite{shen2014learning}) and  Long Short-Term Memory (LSTM) variants \cite{hochreiter1997long,palangi2016deep,wan2016deep,cohen2016end}. But this kind of models are limited by the fixed-size embedding, and thus were soon superseded by interaction-based cross-encoders, most notably BERT-based models, that concatenate query and document as input and predict relevance via a classification or regression head over the [CLS] token. \cite{nogueira2019multi} first introduced this method and proved it effective. These models achieve strong performance but require pairwise inference, limiting scalability, and is constrained by fixed representational dimention and content length.
~
With the rise of LLMs, researchers have repurposed powerful encoder-decoder (e.g., T5 \cite{raffel2020exploring}) or decoder-only (e.g., LLaMA \cite{touvron2023llama}) backbones for reranking. A dominant paradigm \cite{zhuang2023rankt5, ma2024fine} fine-tunes LLMs to regress continuous relevance scores or classify relevance levels, leveraging their rich contextual understanding and long-context capabilities for improved semantic matching. Jina-Reranker-M0 also follows this paradigm but leverages Qwen2-VL \cite{wang2024qwen2}, enabling support for multimodal inputs.
~
More recently, generative reranking has emerged as a paradigm shift: instead of scoring candidates individually, the model directly generates the ranked sequence—such as a list of document indices or IDs—conditioned on the full candidate set. Methods like \cite{yoon2024listt5, yang2025rank} process all candidates jointly and autoregressively output the optimal ranking. This generative formulation naturally integrates reranking into end-to-end AI systems and shows particular promise in multimodal and agent-driven retrieval scenarios.

%% file: sec/3_method.tex
\section{Method}
\label{sec:method}

In this section, we introduce YOFO, an efficient judging paradigm adapting an autoregressive model to receive a series of requirements as input and yields the judgements (``yes'' or ``no'') for all requirements, respectively, as output. We begin by formulating the problem in \cref{subsec:problem}, followed by the overall architecture of YOFO in \cref{subsec:arch}, training in \cref{subsec:train}, inference in \cref{subsec:infer}.

%-------------------------------------------------------------------------

\begin{figure*}[t]
    \centering
    \includegraphics[width=\linewidth]{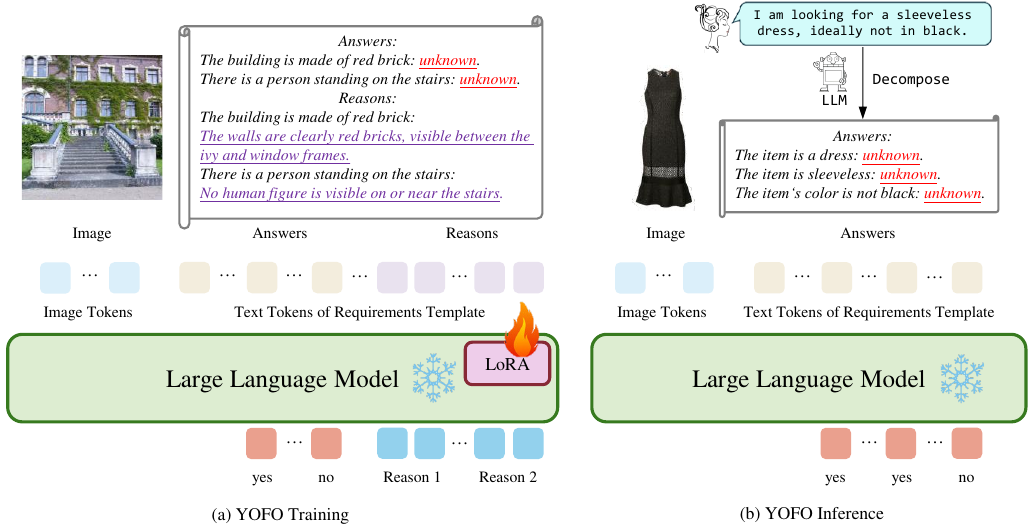}
    \caption{\textbf{Architecture of YOFO.} (Left): YOFO uses an MLLM as its backbone. During training, the MLLM receives a structured requirement template as input and is trained to (i) judge each requirement and (ii) produce supporting reasons. YOFO supervises the model to output correct yes/no judgments while applying next-token prediction only to the reasoning text. (Right): At inference time, the user’s query is first decomposed into a structured template, and the MLLM ingests the image and the template to determine, in a single forward pass, whether each requirement is satisfied, without autoregressively generating reasons.}
    \label{fig:ourarch}
    \vspace{-2mm}
\end{figure*}

%-------------------------------------------------------------------------
\subsection{Problem formulation}
\label{subsec:problem}

As illustrated in \cref{sec:intro}, models akin to Jina-Reranker-M0 \cite{jina2025blog} struggle to perform accurate cross-modal retrieval. One might attribute this to visual information degradation caused by the image encoder and the connector projection, which impairs fine-grained alignment. However, we empirically find that these models also fail to retrieve text accurately from natural-language queries. 
For example, consider the query ``I'm looking for some bottoms that are ideally suitable for summer wear and I'd prefer them to be grey if available'', paired with two documents: (1) ``grey formal trousers, a touch of warmth—perfect for winter'' and (2) ``black skirt, perfect for casual or date outfits in summer'', Jina-Reranker-M0 assigns a higher score to winter trousers simply because they are gray, failing to recognize that the query explicitly requires summer-appropriate clothing, while “gray” is only a conditional preference. 
Verifying each requirement in the query individually would likely yield better results. To this end, we formally define the \textbf{compositional judging problem}.

The input to YOFO consists of an image $I \in \mathbb{R} ^ {3\times H\times W}$ and $N$ requirements $(\bm{p}_i)_{i=1}^N$, for which the model determines whether they hold of the image. YOFO is a function that judges whether each requirement is satisfied with respect to the image:
\begin{equation}
  f(I, (\bm{p}_i)_{i=1}^N) = (\bm{a}_i)_{i=1}^N
  \label{eq:general}
\end{equation}
That is, YOFO judges each requirement $\bm{p}_i$ and outputs a binary answer $\bm{a}_i \in \{\text{yes}, \text{no}\}$.

%-------------------------------------------------------------------------
\subsection{Overall Architecture}
\label{subsec:arch}

YOFO leverages a decoder-only MLLM as the backbone. As illustrated in \cref{fig:ourarch}, we add a single token \texttt{unknown} to the end of each requirement $\bm{p}_i$, and concatenate all requirements to form the template $\bm{t}$. The template $\bm{t}$ is then tokenized into $\bm{s}$, and the indices of the single token \texttt{unknown} in the sequence are denoted $(pos_i)_{i=1}^N$. After a single forward pass, the model must predict whether each \texttt{unknown} position should be ``yes'' or ``no''. The forward process can be formulated as:

\begin{equation}
    \bm{h} = \text{VLM}(I, \bm{t})
    \label{eq:forward}
\end{equation}
The acquired logits $\bm{h}$ indicate the next-token distribution at each position. Therefore, for each requirement $\bm{p}_i$, the logits at position $pos_i - 1$ imply the distribution over the next token at $pos_i$, distinguishing “yes” from “no”. In other words, it is this logit that we need to predict whether the requirement $\bm{p}_i$ is satisfied for the image input $I$. We read out the required logits as:

\begin{equation}
    \bm{l}_i = \bm{h}[pos_i - 1], \quad i = 1, 2, \cdots, N
    \label{eq:logit}
\end{equation}
After that, these logits $\bm{l}_i \in \mathbb{R} ^ {V}$, where $V$ is the vocabulary size of the LLM, are converted into a probability distribution via the softmax function, and the index of the maximum probability yields the model’s prediction:

\begin{equation}
    \bm{a}_i^\prime = \text{vocab}[\arg \max(S(\bm{l}_i))], \quad i = 1, 2, \cdots, N
    \label{eq:preds}
\end{equation}
where $S$ denotes softmax function.
In real-world applications, to ensure that the LLM outputs binary results (``yes''/``no''), we can compare the probabilities of emitting ``yes'' and ``no'', and select the higher one as the final judgment. Therefore, this can be framed as below:

\begin{equation}
    \bm{a}_i^\prime = 
    \begin{cases} 
      \text{``yes''}, & \text{if } P(\text{``yes''}|s_{<pos_{i}}) > P(\text{``no''}|s_{<pos_{i}}) \\
      \text{``no''}, & \text{otherwise}
    \end{cases}
    \label{eq:preds2}
\end{equation}
where $i = 1, 2, \cdots, N$.
Moreover, our experiments empirically reveal that, after training, the LLM learns to output only “yes” and “no”—even without explicit binarization.

%-------------------------------------------------------------------------
\subsection{Training}
\label{subsec:train}

In this section, we describe YOFO’s training procedure in detail. We first present the version without post-hoc Chain-of-Thought (CoT), and then introduce a variant with post-hoc CoT. The only difference is that, in the non–post-hoc CoT setting, the template omits the field of reasons that follow the answers.

\noindent \textbf{Without Post-hoc CoT.} As shown in \cref{fig:ourarch} (a), we apply supervision at the answer positions to train the LLM to judge the requirements correctly. Specifically, we minimize the cross-entropy loss between the model logits $(\bm{l}_i)_{i=1}^N$ and the ground-truth labels. The answer loss $\mathcal{L}_\text{answer}$ is defined as:
\begin{equation}
    \mathcal{L}_\text{answer} = -\frac{1}{N}\sum_{i=1}^{N}\log P(a_{i}|s_{<pos_{i}})
    \label{eq:realoss}
\end{equation}
Although this objective resembles the standard autoregressive loss, when the model computes $\bm{l}_i$ for the $i$-th requirement, it does not observe the answers to previous requirements. Therefore, $\mathcal{L}_\text{answer}$ fundamentally differs from the autoregressive objective. Optimizing $\mathcal{L}_\text{answer}$ enables the LLM to make dependency-aware judgments, allowing later decisions to condition on its earlier judgements, as shown in \cref{subsec:hier}.

\noindent \textbf{With Post-hoc CoT.} As shown in \cref{fig:ourarch} (a), YOFO with Post-hoc CoT applies supervision to both the answers and the accompanying rationales. The requirements themselves are not supervised, as they are provided as input rather than predicted. Let $\mathcal{V}$ denote the positions of tokens in the reason field, that is $\mathcal{V} = \{i|s_{i} \text{ lies in the reason field}\}$. For each $i \in \mathcal{V}$, the logits at the preceding position $(i-1)$ predict the distribution over the next token $s_{i}$ in the reason field. Accordingly, we define the following autoregressive loss to encourage the model to consider rationales prior to making judgments:
\begin{equation}
    \mathcal{L}_\text{reason} = -\frac{1}{|\mathcal{V}|}\sum_{i \in \mathcal{V}}\log P(s_{i}|s_{<i})
    \label{eq:anloss}
\end{equation}
Because reason tokens substantially outnumber answer tokens and $\mathcal{L}_\text{reason}$ is introduced primarily to aid judgment, we scale $\mathcal{L}_\text{reason}$ by a coefficient $\lambda$, and  define the total loss as:
\begin{equation}
    \mathcal{L} = \mathcal{L}_\text{answer} + \lambda \mathcal{L}_\text{reason}
    \label{eq:totloss}
\end{equation}

%-------------------------------------------------------------------------
\subsection{Inference}
\label{subsec:infer}

\cref{fig:ourarch} (b) illustrates how a YOFO-trained LLM performs inference. At inference time, there are no ground-truth annotations for rationales; consequently, the rationale field is omitted from the template. The MLLM backbone ingests the image $I$ and the template $\bm{t}$ and performs a single forward pass, after which the logits $(\bm{l}_i)_{i=1}^N$ can be extracted as in \cref{eq:logit}. Finally, we apply the binary decision rule in \cref{eq:preds2} to ensure that the LLM outputs only ``yes'' or ``no''. The resulting judgments can then be consumed by downstream tasks—for example, provided as input to an agent for further analysis or mapped to a single score under a predefined criterion.

As an illustrative use case of YOFO, \cref{fig:ourarch} (b) depicts how it can be integrated into rerankers. To obtain a structured template, an LLM first decomposes the user’s query into a set of requirements. These requirements are then assembled into a template and passed, together with the image, to the MLLM. Finally, all requirements are judged in a single forward pass, yielding results that substantially improve reranking quality. Task-specific rules can be devised to exploit these judgments; we leave the design open to practitioners.

%% file: sec/4_experiments.tex
\section{Experiments}
\label{sec:experiments}

In this section, we present extensive experiments and ablation studies to validate our approach. We first compare YOFO with state-of-the-art (SOTA) methods to demonstrate its effectiveness. We then assess YOFO’s dependency-aware judgement, the effectiveness of post-hoc CoT, and the contributions of individual architectural components.

%-------------------------------------------------------------------------
\subsection{Implementation Details}
\label{subsec:imple}

We build on Qwen2-VL-2B-Instruct \cite{wang2024qwen2} and Qwen3-VL-2B-Instruct \cite{bai2025qwen3vltechnicalreport}, freezing the vision encoder during training. We fine-tune the LLM using LoRA \cite{hu2022lora}. To conveniently locate the \texttt{unknown} token, we add four special tokens to the Qwen2-VL-2B-Instruct and Qwen3-VL-2B-Instruct processor: \texttt{\textless\textbar auth\_start\textbar\textgreater}, \texttt{\textless\textbar auth\_end\textbar\textgreater}, \texttt{\textless\textbar reason\_start\textbar\textgreater}, and \texttt{\textless\textbar reason\_end\textbar\textgreater}. The first two mark the start and end of the answer span, and the last two mark the boundaries of the reason span. Consequently, these special tokens enable straightforward label masking, allowing supervision to be applied exclusively to answers and reasons.

Training is conducted for one epoch with a learning rate of $1 \times 10^{-4}$, using the AdamW optimizer. We employ a cosine learning rate scheduler with a warmup ratio of 0.05. We set the maximum sequence length to 1{,}536 tokens. We use DeepSpeed ZeRO-3  \cite{rajbhandari2020zero} for distributed training. We load the training data with eight worker processes, dropping the last incomplete batch. Training runs on eight H100 GPUs over four hours. We use bfloat16 mixed precision and enable gradient checkpointing to reduce memory usage. Moreover, we adopt $\lambda = 0.55$ as the default setting in all experiments using post-hoc CoT.

%-------------------------------------------------------------------------

\begin{figure*}[t]
    \centering
    \includegraphics[width=0.8\textwidth]{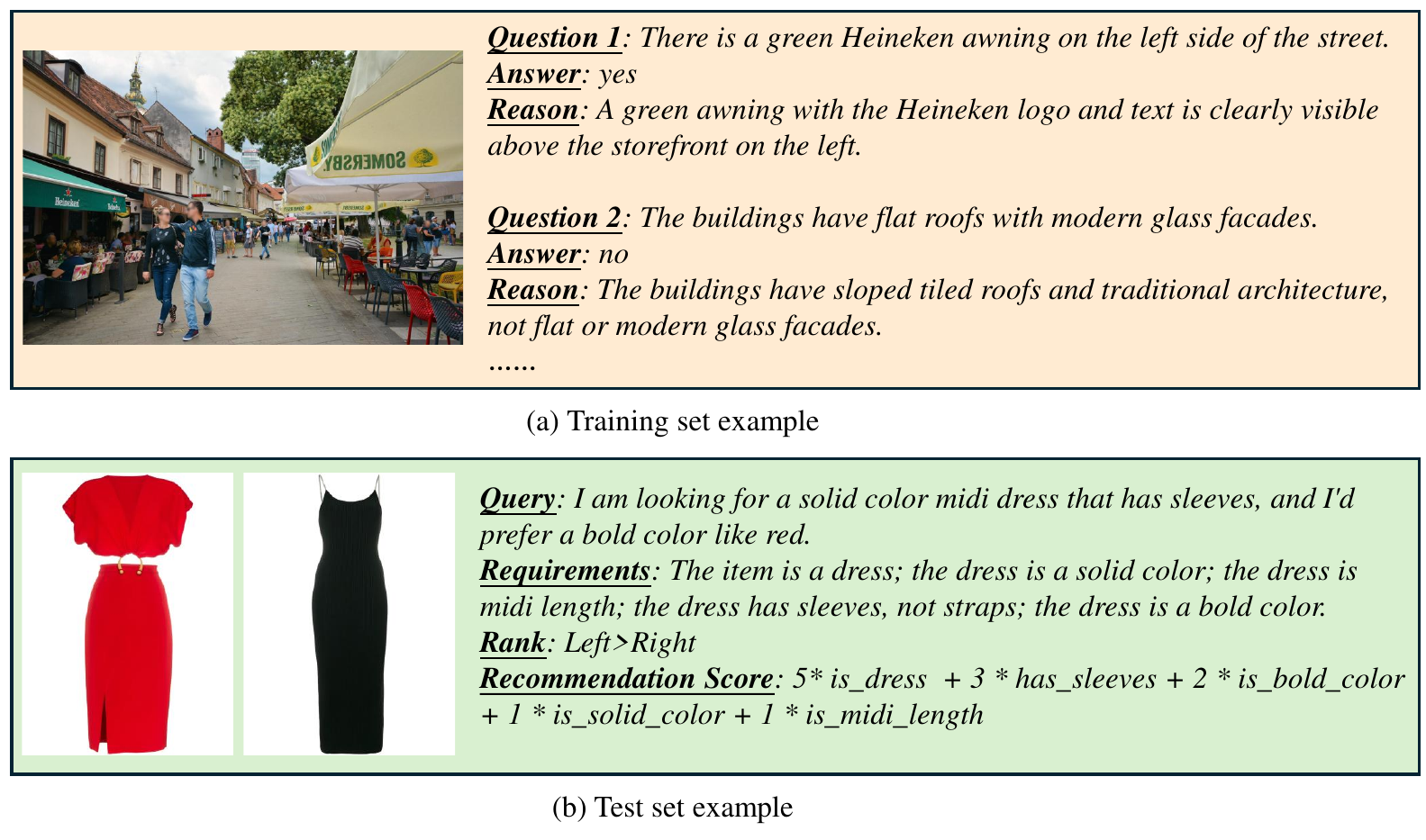}
    \caption{\textbf{Examples of the training and test sets.} Each training sample consists of an image, a set of properties, answers, and corresponding reasons. Each test sample consists of two images, a customer-style query, requirements derived by decomposing the query, ground-truth labels indicating which image better matches the query, and the expression used to compute each image's final recommendation score.}
    \label{fig:dataex}
    \vspace{-4mm}
\end{figure*}

%-------------------------------------------------------------------------
\subsection{Data Construction}
\label{subsec:datacons}

\textbf{Training Set.} To construct the training dataset, we use the SA-1B dataset \cite{kirillov2023segment}, introduced alongside the Segment Anything Model. It contains approximately 11M diverse, high-resolution images sourced from a privacy-preserving data engine and annotated with 1.1B high-quality segmentation masks. Since we do not require the masks, as shown in \cref{fig:dataex} (a), we use only the images, prompting an MLLM to analyze them, propose properties that hold or do not hold for the corresponding images, and provide reasons. We randomly sample 1.2M images to ensure diversity and generalizability. For each image $\mathcal{I}_i$, the final sample is represented as $(\mathcal{I}_i, (\mathcal{P}_{ij}, \mathcal{A}_{ij}, \mathcal{R}_{ij})_{j=1}^N)$, denoting the image, the proposed properties, the binary answers(``yes'' or ``no''), and the reasons. To maintain a balanced and randomized mix of satisfied and unsatisfied properties, we prompt the MLLM to produce approximately equal numbers of each and shuffle them during training. Additionally, although we instruct the model to generate exactly ten properties per sample in the training set, the number of properties per training sample varies, typically remaining close to ten.. To evaluate the model and mitigate overfitting, we split the dataset at a ratio of $6:1$ into training and validation sets.

\noindent \textbf{Test Set.} We use the LRVS-Fashion dataset \cite{lepage2023lrvs}, a large-scale public benchmark for referred visual search (RVS) in the fashion domain, to construct a test set. It comprises 272K fashion products and 842K images extracted from real-world catalogs. To evaluate generalization in Image-Text reranking, as depicted as \cref{fig:dataex} (b), we construct image pairs and queries; the model then judges which of the two images better matches the corresponding query. Specifically, we randomly sample image pairs $(\mathcal{I}_{i1}, \mathcal{I}_{i2})$ from the same category and, using an MLLM, generate a natural-language query $\mathcal{Q}_{i}$ that mimics how a customer would express their preferences. The MLLM then produces the answer $\mathcal{A}_{i}$ indicating whether the first image $\mathcal{I}_{i1}$ better matches $\mathcal{Q}_{i}$, the set of query requirements $(\mathcal{R}_{ij})_{j=1}^{M_i}$ (where $M_i$ is the number of requirements), and a valid python expression $\mathcal{E}_{i}$ used to compute the final recommendation score based on our model's predictions.

\textbf{Data Curation.} To ensure the quality of the training and test sets generated by MLLMs, we implement a human-in-the-loop curation protocol. For training set and test set independently, we iteratively refine the corresponding prompt, generate samples from a small, randomly selected subset of the source data, and manually evaluate sample quality and correctness against predefined standard. This iteration continues until the generated samples consistently satisfy the quality thresholds. The finalized prompts are then employed to produce the complete training and test sets, respectively.

%-------------------------------------------------------------------------
\subsection{Evaluation Metrics}
\label{subsec:metrics}

In this section, we introduce the metrics used in our experiments to evaluate models.

On SA-1B validation sets and dependency-aware judgement task, we report two metrics to evaluate the accuracy of models. One metrics is Sample-wise Accuracy $Acc_\mathrm{sample}$, denoting the percentage of samples on which the model judges all requirements correctly. Another metrics is Property-wise Accuracy $Acc_\mathrm{property}$, indicating the percentage of properties judged correctly. Obviously, $Acc_\mathrm{sample}$ is always not greater than $Acc_\mathrm{property}$.

To evaluate the model’s ability to make judgments conditioned on its previous decisions, we introduce an additional metric for the dependency-aware judgment task, termed Dependency-Aware Accuracy and denoted as $Acc_\mathrm{dep}$. It measures the percentage of questions that depend on previous ones and are answered correctly.

On the test sets, we evaluate whether the model can rank the image pairs properly. Therefore, we report the error rate of ranking $Err_\mathrm{rank}$, denoting the percentage of image pairs ranked wrongly.

%-------------------------------------------------------------------------
\subsection{Comparison on LAION-RVS-Fashion}
\label{subsec:laion}

In this section, we present a comparative analysis of YOFO with representative state-of-the-art embedding models (e.g., CLIP \cite{radford2021learning}, SigLIP 2 \cite{tschannen2025siglip} and BLIP-2 \cite{li2023blip}) and multimodal rerankers (e.g., Jina-Reranker-M0 \cite{jina2025blog}, Qwen3-VL-Reranker \cite{li2026qwen3} and LamRA-Rank \cite{liu2025lamra}). Notably, all evaluated rerankers are built on Qwen2-VL \cite{wang2024qwen2} or Qwen3-VL \cite{bai2025qwen3vltechnicalreport}.

LAION-RVS-Fashion is a specialized fashion dataset, which inherently exhibits a significant domain shift relative to our training set constructed entirely on the broad and diverse SA-1B corpus. Despite this distributional gap, YOFO demonstrates remarkable robustness and adaptability when evaluated on fashion-specific recommendation tasks. As shown in \cref{tab:laion}, YOFO achieves competitive performance and consistently outperforms all strong baselines—despite never being trained on fashion-related data. Notably, LamRA-Rank-7B (fine-tuned on Qwen2-VL-7B), despite its significantly larger parameter count, exhibits a ranking error rate nearly twice that of YOFO finetuned on Qwen2-VL-2B. Similarly, Qwen3-VL-Reranker-2B outperforms LamRA-Rank-7B, but still underperforms YOFO. This performance gap is attributable to the inherent limitation of single-score prediction paradigm in capturing fine-grained semantic relationships.

This result highlights a key advantage of our approach: YOFO learns general-purpose judging capabilities that transfer effectively across domains. Crucially, it can be deployed directly in specialized subdomains—such as fashion—without any fine-tuning or domain adaptation. This zero-shot generalization underscores the model’s practical utility in real-world scenarios where labeled data is scarce or domain shifts are common, positioning YOFO as a versatile solution for cross-domain recommendation systems.

Regarding throughput evaluation, throughput metrics for embedding models are omitted, as their embedding process is typically performed offline to accelerate retrieval in modern recommendation systems. Among the evaluated multimodal rerankers, YOFO finetuned on Qwen3-VL-2B achieves the highest throughput, demonstrating superior computational efficiency.

Additionally, YOFO based on Qwen3-VL-2B achieves a 1.1\% lower ranking error rate and higher throughput compared to its Qwen2-VL-2B counterpart, which we attribute to Qwen3-VL-2B’s smaller parameter count.

\newcolumntype{C}{>{\centering\arraybackslash}X}

\begin{table}[t]
\caption{\textbf{Ranking Error Rates on LAION-RVS-Fashion252 Dataset.} YOFO achieves the lowest ranking error rate across all baseline models while simultaneously attaining the highest throughput among all evaluated rerankers.}
\centering
\begin{tabularx}{\columnwidth}{lcC}
\toprule
Methods & $Err_\mathrm{rank}$ (\%) $\downarrow$ & $\mathrm{Throughput}$ (pairs/s) $\uparrow$ \\
\midrule
Clip-ViT-L & 42.0 & - \\
Siglip2-ViT-G & 48.3 & - \\
Blip2-itm-ViT-G & 37.4 & - \\
\midrule
Jina-Reranker-m0 & 16.2 & 36 \\
Qw3-Reranker-2B & 8.7 & \textbf{48} \\
LamRA-Rank-7B & 9.3 & 5 \\
\midrule
YOFO (Qwen2-VL) & 4.8 & 35 \\
\textbf{YOFO (Qwen3-VL)} & \textbf{3.7} & \textbf{48} \\
\bottomrule
\end{tabularx}
\label{tab:laion}
\vspace{-2mm}
\end{table}

%-------------------------------------------------------------------------
\subsection{Dependency-Aware Judgement}
\label{subsec:hier}

To evaluate whether our model can make judgments conditioned on previous ones, we randomly select two properties from each sample in the validation set and replace the second property with ``\textbf{The answer to this question is the opposite of the answer to the previous question}''. The model is therefore required to infer the answer to the second question without being given the first property's answer, while making both judgments simultaneously. We fine-tune the pretrained Qwen2-VL-2B-Instruct model on this dataset for one epoch, keep other settings unchanged, and then compare it with the model in \cref{subsec:laion}. Although challenging, \cref{tab:hier} demonstrates YOFO’s dependency-aware judgment ability under the dependent-question setup, approaching 100\%. Without training on the purposely-designed dataset, YOFO cannot judge correctly, because the properties in the training and validation sets can be judged individually, rendering dependency-aware analysis unnecessary. The base model performs poorly on this task, as it has not been exposed to scenarios where previous tokens are unobserved when predicting the next token.

\begin{table}[t]
\caption{\textbf{Accuracy on SA-1B validation set.} ``Base''denotes pretrained Qwen2-VL-2B-Instruct model. ``YOFO + dep'' denotes YOFO trained with the dependency question. To construct questions that need dependency-aware analysis based on previous judgments, we randomly choose two properties from each sample in validation set, and replace the second requirement with ``The answer to this question is the opposite of the answer to the previous question.''}
\centering
\begin{tabularx}{\columnwidth}{Xcc}
\toprule
Methods & $Acc_\mathrm{dep}$ (\%) $\uparrow$ & $Acc_\mathrm{property}$ (\%) $\uparrow$ \\
\midrule
Base & 35.3 & 62.9 \\
YOFO  & 57.6 & 71.4 \\
\textbf{YOFO + dep} & \textbf{99.1} & \textbf{90.4} \\
\bottomrule
\end{tabularx}
\label{tab:hier}
\end{table}

%-------------------------------------------------------------------------
\begin{figure*}[t]
    \centering
    \includegraphics[width=\linewidth]{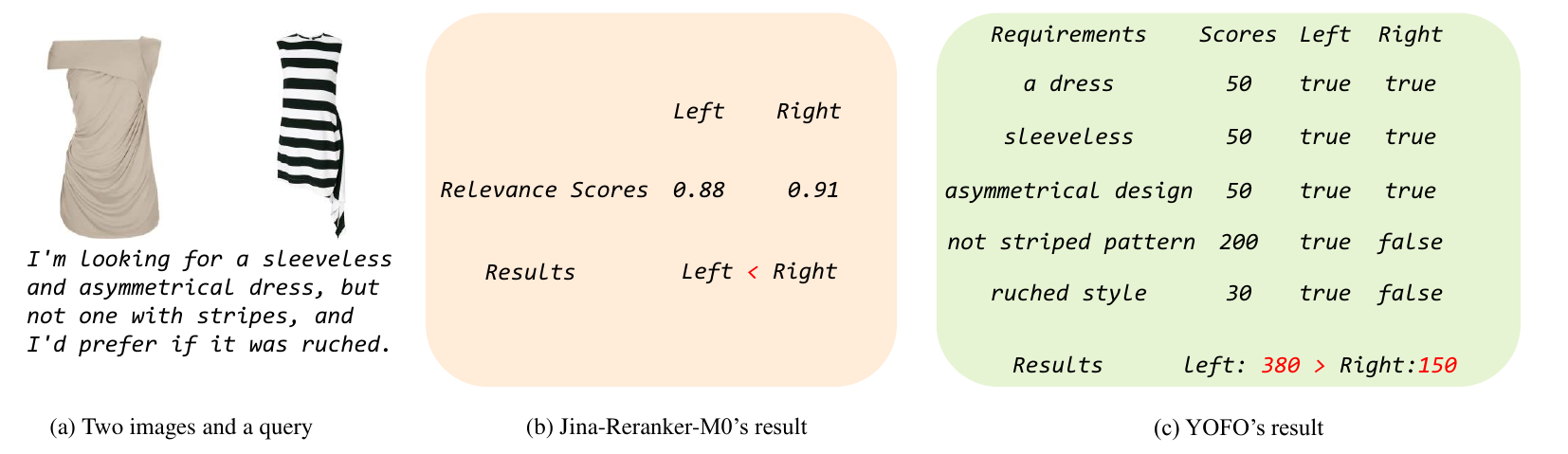}
    \caption{\textbf{Illustrative comparison using a single example.} (a) presents a pair of images and a query. (b) shows that Jina-Reranker-M0 assigns a higher score to the second image; however, the dress in that image is striped and not ruched, so the prediction is incorrect. Panel (c) shows that after decomposing the query into five requirements, YOFO judges whether each requirement is satisfied by each image; these judgments are then combined using the specified expression to compute the final recommendation scores. Because all judgments are correct, the resulting prediction is accurate.}
    \label{fig:comps}
\end{figure*}

\begin{figure}[t]
    \centering
    \includegraphics[width=1\linewidth]{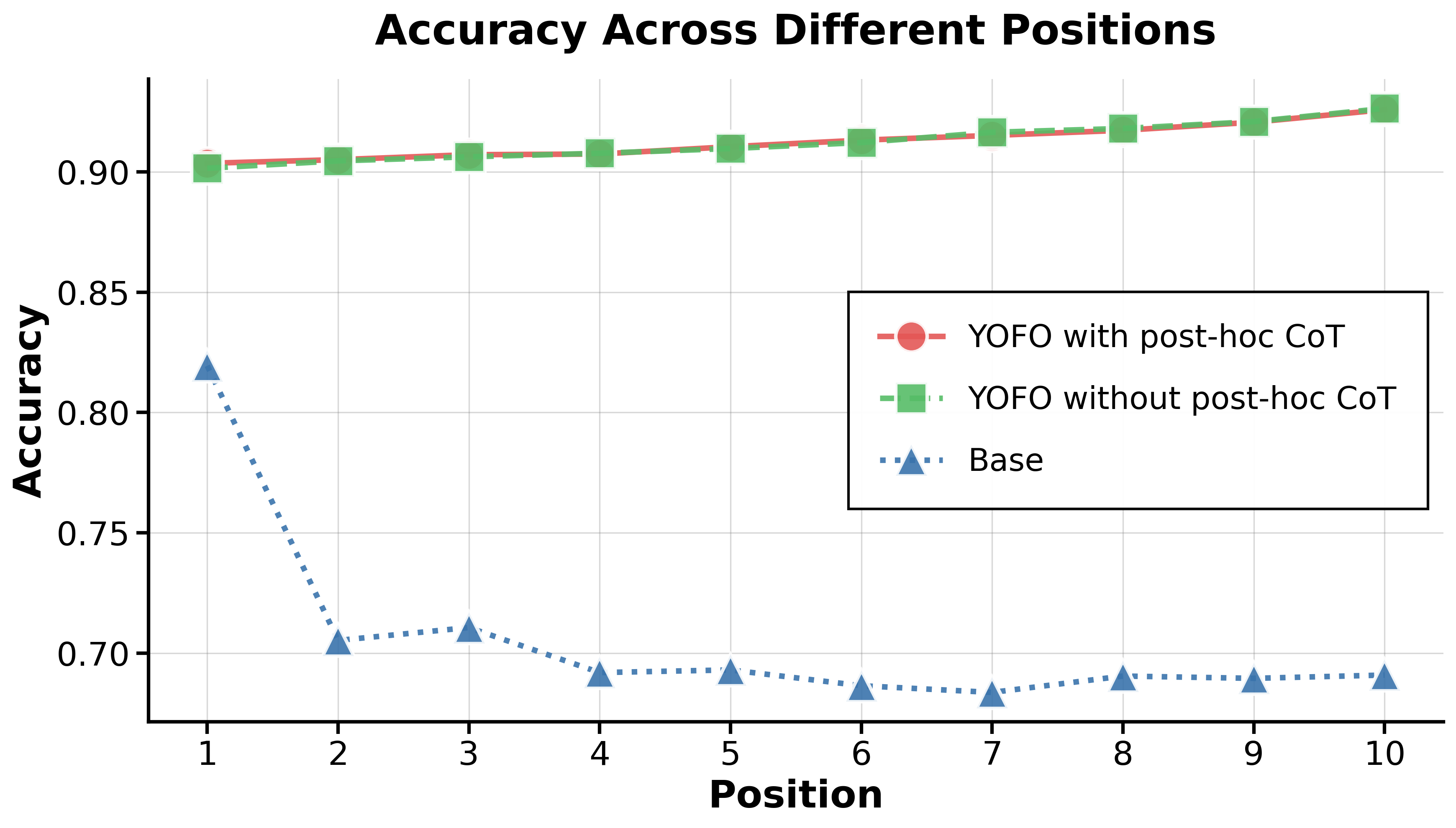}
    \caption{\textbf{The accuracy of different models across template positions on the SA-1B validation set.} ``Base'' denotes Qwen2-VL-2B-Instruct. As properties appear later in the sequence, YOFO maintains an accuracy of approximately 90\%, while the base model shows a sharp decline in accuracy starting from the second property.}
    \label{fig:add}
    \vspace{-4mm}
\end{figure}

%-------------------------------------------------------------------------

\subsection{Ablation Study}
\label{subsec:ablation}

\textbf{Effectiveness of YOFO training.} We fine-tune Qwen2-VL-2B-Instruct and Qwen3-VL-2B-Instruct models on the training set using LoRA \cite{hu2022lora} for one epoch and evaluate it on the validation set. As shown in \cref{tab:ablation}, after training, Property-wise Accuracy improves by over 20\%, reaching above 90\%, while Sample-wise Accuracy increases by approximately 40\%. These ablation results demonstrate the effectiveness of our training paradigm. Moreover, the YOFO model trained with Qwen3-VL-2B-Instruct performs slightly better than its counterpart trained with Qwen2-VL-2B-Instruct. The Sample-wise Accuracy remains moderate because the training and validation sets contain, on average, nearly ten properties per sample, making it challenging to judge all properties correctly.

\begin{table}[t]
\caption{\textbf{Ablation study of different components on SA-1B validation set.} We train models with Qwen2-VL-2B-Instruct and Qwen3-VL-2B-Instruct as base models. ``YOFO'' denotes the model obtained by fine-tuning the base model with YOFO. ``YOFO + CoT'' denotes the YOFO model augmented with post-hoc CoT.}
\centering
\begin{tabularx}{\columnwidth}{Xcc}
\toprule
Methods & $Acc_\mathrm{property}$ (\%) $\uparrow$ & $Acc_\mathrm{sample}$ (\%) $\uparrow$ \\
\midrule
\textcolor{gray}{\textit{Qwen2-VL}} & & \\
Base & 70.6 & 2.0 \\
YOFO & 91.2 & 41.8 \\
\textbf{YOFO + CoT} & \textbf{91.3} & \textbf{42.0} \\
\midrule
\textcolor{gray}{\textit{Qwen3-VL}} & & \\
Base & 68.4 & 1.3 \\
\textbf{YOFO} & \textbf{92.3} & \textbf{46.6} \\
\bottomrule
\end{tabularx}
\label{tab:ablation}
\vspace{-2mm}
\end{table}

\noindent \textbf{Effectiveness of post-hoc CoT.} As shown in the last row of \cref{tab:ablation}, YOFO benefits from post-hoc CoT. This likely arises because the model is trained to generate and consider reasons before making judgments. This component is particularly useful in complex settings where detailed reasons can be annotated.

\noindent \textbf{Rank of LoRA.} We evaluate different ranks of LoRA, and the results in \cref{tab:rank} show that $r=64$ perform the best. Therefore, we adopt $r=64$ as our default experimental setting.

\begin{table}[t]
\caption{\textbf{Ablation study on LoRA rank using the SA-1B validation set.} As the LoRA rank increases, accuracy first rises and then declines, indicating that a rank of 64 yields the best performance.}
\centering
\begin{tabular}{lcc}
\toprule
Rank & $Acc_\mathrm{property}$ (\%) $\uparrow$ & $Acc_\mathrm{sample}$ (\%) $\uparrow$ \\
\midrule
32 & 91.2 & 41.7 \\
\textbf{64} & \textbf{91.3} & \textbf{42.0} \\
128 & 91.2 & 41.7 \\
\bottomrule
\end{tabular}
\label{tab:rank}
% \vspace{+1mm}
\end{table}

\subsection{Additional Analysis}
\label{subsec:addana}

We present a case study in \cref{fig:comps}. The user seeks a sleeveless, asymmetric dress that is not striped and preferably has ruching. However, jina-reranker-m0 assigns a higher score to the dress in the second image, overlooking its stripes and lack of ruching. In contrast, YOFO judges, for each image, whether the query’s requirements are satisfied and correctly assigns a higher score to the first image. This case explicitly demonstrates the effectiveness and interpretability of our method.

We also analyze judgment accuracy across positions on the SA-1B validation set. Specifically, $Acc_i$ denotes the accuracy for $i$-th requirement, and \cref{fig:add} visualizes the results. The figure shows that the base model achieves over 80\% accuracy on the first requirement, but its accuracy declines rapidly toward 70\% thereafter. In contrast, YOFO maintains accuracy above 90\%, with or without post-hoc CoT. This likely occurs because the base model’s behavior on the first requirement resembles autoregressive generation, yet it fails to adapt to our template-conditioned judgment paradigm in which the ground-truth labels for previous judgments are unobserved. After training, YOFO adapts to this paradigm and consequently maintains high accuracy across positions.

%% file: sec/5_conclusions.tex
\section{Conclusions}
\label{sec:conclusions}

In this work, we present You Only Forward Once (YOFO), a template-conditioned judging paradigm that judges whether inputs satisfy requirements specified by a predefined template. By concurrently predicting next-token distributions for all requirements in a single forward pass, YOFO efficiently judges compositional requirements. Although trained on general-purpose data, YOFO achieves state-of-the-art performance on the reranking task and generalizes well across domains, enabling direct deployment in specialized settings. Our precise and efficient approach departs from traditional methods, which either lack fine-grained understanding of inputs or incur prohibitive computational overhead. Additionally, YOFO can condition later judgments on earlier ones and benefits from post-hoc CoT, making it effective in complex settings. Ultimately, our work establishes a new paradigm for judging, offering practical advantages in scenarios where multiple requirements must be assessed.

\noindent \textbf{Limitations and Future Work.} While our work demonstrates the effectiveness of YOFO in fine-grained multimodal compositional judging, it has a limitation that points to promising directions for future research. Specifically, we evaluate YOFO only on a single downstream task: reranking. Additional application scenarios warrant investigation. Future work could investigate additional applications of YOFO. Its explicit, interpretable judgments naturally lend themselves to serving as signals in reinforcement learning (RL) frameworks. Specifically, YOFO could be used as a structured reward model that provides fine-grained feedback (e.g., per-requirement rewards) rather than a single scalar score, thereby offering more precise guidance during RL training of MLLMs and diffusion models. Moreover, YOFO is well suited to multi-label classification and can therefore be applied to numerous scenarios, such as personalized recommendation systems that tag products and users’ interests to deliver targeted recommendations.